\documentclass{article}

\PassOptionsToPackage{numbers, compress}{natbib}
\usepackage[preprint]{neurips_2026}
\usepackage[utf8]{inputenc} 
\usepackage[T1]{fontenc}    
\usepackage[hidelinks]{hyperref}  
\usepackage{url}            
\usepackage{booktabs}       
\usepackage{amsfonts}       
\usepackage{nicefrac}       
\usepackage{microtype}      
\usepackage{xcolor}         
\usepackage{mdframed}

\usepackage[textsize=tiny]{todonotes}
\usepackage{algorithm}
\usepackage{algpseudocode}
\usepackage{amsmath}
\usepackage{amssymb}
\usepackage{comment}
\usepackage{tcolorbox}
\usepackage{subcaption}
\usepackage{mathtools}
\usepackage{enumitem}
\usepackage{colortbl}
\usepackage{multirow}
\usepackage{wrapfig}
\usepackage{listings}
\newcommand{\datasetname}{\textsl{MathAtlas}}
\newcommand{\hardname}{\textsl{MA-Hard}}
\newcommand{\alignmentname}{\textsl{MA-Align}}

\newcommand{\nilay}[1]{\todo[inline,color=blue!40]{\small #1 -- Nilay}}
\newcommand{\jmf}[1]{\todo[inline,color=green!40]{\small #1 -- Jeff}}
\renewcommand{\nilay}[1]{}
\renewcommand{\jmf}[1]{}

\title{MathAtlas: A Benchmark for Autoformalization\\ \textit{in the Wild}}

\author{%
  Nilay Patel\textsuperscript{$\dagger$*}\\
  \texttt{nilay@ucsc.edu} \\
  \And
  Noah Arias\\
  \texttt{ngarias@ucsc.edu} \\
  \And
  Davit Babayan\\
  \texttt{dbabaya1@ucsc.edu} \\
  \AND
  Victoria Cochran\\
  \texttt{vacochra@ucsc.edu} \\
  \And
  Timothy Libman\\
  \texttt{tlibman@ucsc.edu} \\
  \And
  Hafsah Mahmood\\
  \texttt{hzmahmoo@ucsc.edu} \\
  \And
  Liam McCarty\\
  \texttt{lmccarty@ucsc.edu} \\
  \And
  Soli Munoz\\
  \texttt{mmunoz35@ucsc.edu} \\
  \And
  Laurel Willey\\
  \texttt{lwwilley@ucsc.edu} \\
  \AND
  Jeffrey Flanigan\textsuperscript{*}\\
  \texttt{jmflanig@ucsc.edu} \\ \\
  {University of California, Santa Cruz}\\
  {\textsuperscript{$*$}Primary authors}\\
  {\textsuperscript{$\dagger$}Correspondence}
}

\begin{document}

\maketitle

 \begin{abstract}
    Current autoformalization benchmarks are largely focused on olympiad or undergraduate mathematics, while graduate and research-level mathematics remains underexplored.
    In this paper, we introduce \datasetname, the first large-scale autoformalization benchmark of \textit{in the wild} graduate-level mathematics, containing $\sim 52k$ theorems, definitions, exercises, examples, and proofs extracted from 103 graduate mathematics textbooks. 
    \datasetname~is enriched with a mathematical dependency graph containing $\sim178k$ relations, and is the first autoformalization benchmark to include such relations, facilitating evaluation and development of dependency-aware autoformalization systems.
    Our extensive experiments show that \datasetname~is high quality but extremely challenging: strong baselines achieve at most 9.8\% correctness on theorem statements and 16.7\% on definitions. Furthermore, we find performance of state-of-the-art models degrades substantially with dependency depth: on \hardname, a subset of 700 entities with the deepest dependency trees, the best model achieves only 2.6\% correctness for autoformalization on this challenging dataset.  We release \datasetname~to the community as a benchmark set for large-scale autoformalization of graduate-level mathematics \textit{in the wild}.
 \end{abstract}

\section{Introduction}
\label{sec:introduction}

The popularity of formal mathematics (computer verifiable mathematics) within the mathematics community has grown dramatically over recent years \citep{Scholze_Buzzard_2020,Tao_2023,Kontorovich_Prime_Number_Theorem_2024}. Together with advances in AI, this has caused a surge of interest in \textbf{autoformalization} \citep{gadgilAutomatingFormalisationTheorem, wuAutoformalizationLargeLanguage2022,jiang2023multilingual,azerbayevProofNetBenchmarkAutoformalizing,liu2025rethinking}, the task of automatically converting natural language math to a formal verifiable language such as Lean \citep{Lean4}.

\begin{figure}[t]
  \centering
  \includegraphics[width=\linewidth]{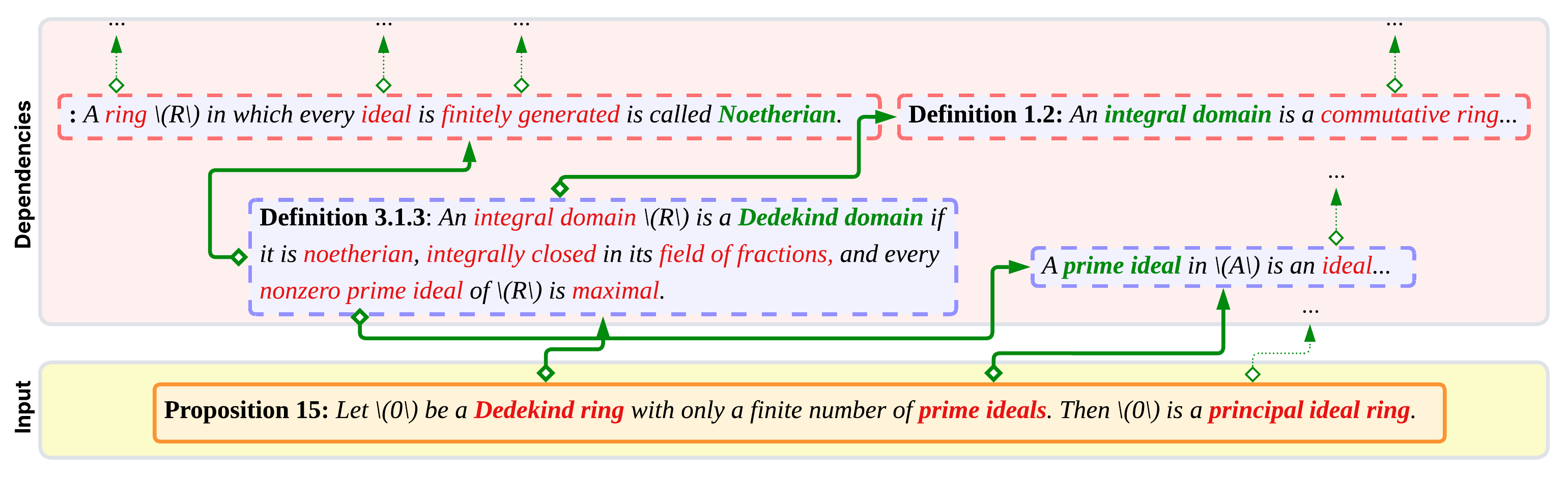}
  \caption{An entity from \datasetname~together with a subgraph of its dependency tree. In yellow (bottom), we have a target theorem statement to be formalized (Proposition 15). Proposition 15 contains three ``object references'' to previously defined mathematical concepts: Dedeking ring, prime ideals, and principal ideal ring, which are dependencies needed to correctly formalize the theorem. These objects each have their own dependency tree (shown with a red border), and so on. Formalizations of objects possibly already exist (e.g., in Mathlib), or they need to be synthesized before the target statement can be formalized.}
  \label{fig:dataset-example}
\end{figure}

Until recently, autoformalization work has been largely focused on simpler settings such as olympiad or undergraduate mathematics \cite{azerbayevProofNetBenchmarkAutoformalizing, wuAutoformalizationLargeLanguage2022, zhengMiniF2FCrosssystemBenchmark2021, ExplorationNeuralMachine, agrawalMathematicsFormalisationAssistant2022, gao2025heraldnaturallanguageannotated, liu2025atlasautoformalizingtheoremslifting}, where mathematical statements require limited prerequisite material to properly formalize, much of which already has been formalized by humans in libraries such as Mathlib \cite{ThemathlibCommunity2020}.

However, in more advanced mathematics, autoformalization becomes increasingly complex. Advanced mathematics has a large amount of prerequisite theory, much of which has yet to be formalized. This means an autoformalization system must first retrieve or produce the necessary dependent theory, correctly formalize it, and then formalize the original target statement. 

Additionally, existing benchmarks typically focus on theorem statements, and ignore formalizations of mathematical \textit{definitions}. Systems which can only autoformalize theorem statements are limited in their scope by the level of existing theory because they cannot formalize theorems using math concepts they do not already have definitions for. Thus, creating scalable autoformalization systems necessitates the study of definition autoformalization, for which large-scale real-world benchmarks are lacking \cite{zhang-etal-2025-autoformalization,wang2025ariaagentretrievaliterative}.

To comprehensively evaluate the performance of autoformalization methods in higher-level mathematics, we present \textbf{\datasetname}\footnotemark, a large-scale, graduate-level benchmark designed to measure autoformalization performance on mathematics in the wild. \datasetname~contains $\sim51k$ informal entities, divided into statements (theorems, examples, and exercises), definitions, and proofs, which are extracted from 103 graduate math textbooks. Entities in \datasetname~span 87 areas of mathematics, including real analysis, differential geometry, category theory, topology, quantum groups, among others. Some entities in \datasetname~are formalizable on top of Mathlib, but many require additional formal theory which may not currently exist. To facilitate access to these required dependencies, we enrich each entity with long-range dependency relations, local variable references, and local context (see \autoref{fig:dataset-example}).

\datasetname~is the first large-scale, \textit{in the wild} autoformalization benchmark of graduate-level mathematics, the first to include rich contextual and dependency data, and the first to annotate their inter-entity dependencies. Our extensive experiments show that \datasetname~is very challenging: strong baseline methods have low overall autoformalization correctness (10.5\% on statements and 20.3\% on definitions) on \datasetname. Furthermore, we find this performance is correlated with an entity's \textit{dependency depth}: our strongest baseline has only 2.6\% overall correctness on \textbf{\hardname}, a subset of 700 entities with the deepest dependency trees.
\footnotetext{\url{https://huggingface.co/datasets/MathAtlas/MathAtlas}}

Finally, we investigate how well existing semantic faithfulness metrics for autoformalization generalize to \datasetname. We present \textbf{\alignmentname}\footnotemark, a binary classification benchmark of 200 entities from \datasetname. Unlike prior benchmarks such as ConsistencyCheck \cite{chen2026reformreflectiveautoformalizationprospective} and CriticLeanBench\cite{peng2025criticleancriticguidedreinforcementlearning}, \alignmentname~contains both labeled examples of theorem statements and definitions, and covers a wide variety of mathematics at the graduate level. Our experiments show that prior baselines have significant room for improvement on \alignmentname.

\footnotetext{\url{https://huggingface.co/datasets/MathAtlas/MA-Align}}

We organize the rest of the paper as follows: In \autoref{sec:mathatlas-benchmark}, we present \datasetname, its construction, contents and quality. We then discuss the benchmark's autoformalization metrics (\autoref{sec:metrics}), before moving on to our autoformalization experiments and an analysis of the results (\autoref{sec:autoformalization-experiments}). Finally, we discuss related works (\autoref{sec:related-works}), and conclude (\autoref{sec:conclusion}).


\section{MathAtlas Benchmark}%
\label{sec:mathatlas-benchmark}

\begin{figure}
\small
  \begin{mdframed}[backgroundcolor=red!10] 
    \textbf{Definition 2.7.10} (Invariant bilinear form).: Let $V=V_{\bar{0}}\oplus V_{\bar{1}}$ be a superspace and carrying a representation of a Lie algebra $\mathfrak{g}$. A bilinear form $(\cdot|\cdot)$ is said \textcolor{red}{\textit{invariant}} if, for any homogeneous element $X\in\mathfrak{g}$ and 
    $v,w\in V$, $(Xv|w)+p(v,X)(v|Xw)=0$. Moreover, if $V$ is the representation space of the adjoint representation, then the bilinear form is said \textcolor{red}{\textit{adjoint}} invariant.
  \end{mdframed}
  \begin{mdframed}[backgroundcolor=blue!10] 
    \textbf{Corollary 3.A.24}: Let \(X\) be a compact Kahler manifold. Then the inclusion \(i:(\mathcal{A}^{*}(X)^{c},d)\subset(\mathcal{A}^{*}(X),d)\) is a dga-quasi-isomorphism.
  \end{mdframed}
  \begin{mdframed}[backgroundcolor=green!10] 
    \textbf{Example 5.1.14}.: Find two fundamental matrices for the linear homogeneous system in Example 5.1.11.
  \end{mdframed}
  \caption{Several entities from \datasetname~of various types: (1) a definition entity which defines two terms: ``invariant'' and ``adjoint invariant''. (2) a theorem statement of a corollary from complex geometry. (3) an example from an ODE textbook.}
  \label{fig:entity-examples}
\end{figure}

\datasetname~is constructed automatically from textbook documents. We discuss the entity and relation types in \datasetname, and then discuss our entity and relation extraction pipeline. We also measure the quality of \datasetname~with a human evaluation.

\subsection{Structure of \datasetname}%
\label{subsec:dataset-structure}

The structure of \datasetname~is as follows, with some relevant statistics.

\paragraph{Entities} We use the term \textbf{entity} to mean a span of informal text which can be formalized, such as a theorem in a textbook. In \datasetname, entities are either definitions, theorems, proofs, examples, or exercises (see \autoref{fig:entity-examples}). Entities can, but don't always, have \textbf{identifiers}, such as ``Theorem 4.3.'' These identifiers are sometimes referenced by other entities (see \autoref{fig:dataset-example}). Importantly, entities sometimes contain multiple formalizable items. For example, in \autoref{fig:entity-examples}, Definition 2.7.10 actually defines two objects, and so it's formalization may require two definitions. 

\paragraph{References} Entities can contain \textbf{references} to other entities, dependent mathematical objects, or local variables which are not explicitly defined in the entity itself. We call these \textit{entity references}, \textit{object references}, and \textit{local variable references}, respectively.

\paragraph{Relations} We also include relations between entity references and the target entity, and object references and entities which define those objects. For example, an entity $e_{met}$ may define a \textit{metric space}, and contain an object reference $r_{top}$ to a \textit{topological space}, which is defined in an entity $e_{top}$. In this case, we add a relation $\left<e_{met}, e_{top}\right>$ (see \autoref{fig:dataset-example}). We define \textit{dependency depth} to be the maximum height of an items dependency tree, where nodes are \textit{entities} (e.g., definitions or theorems) and edges are \textit{object references}. Similarly, we define \textit{dependency mass} to be the total number of unique nodes in a particular entity's dependency tree.

\paragraph{Statistics} We now discuss some statistics of \datasetname. Overall, \datasetname~contains of 52,052 entities, which are comprised of $\sim18$k theorems, $\sim10$k exercises, $\sim10$k definitions, $\sim10$k proofs, and $5$k examples.

About 90\% of entities contain at least one object reference. In sum, there are $\sim$193k object references, for an average of 3.69 per entity. Of these, our relation extraction system found a matching target entity for 83.7\%, for a total of $\sim$161k total object relations in the full graph. In addition, 20.5\% of entities contain an entity reference, summing to $\sim17k$ total entity references. Of these, over 99.8\% were matched with a target entity.

We now briefly present some features of \datasetname's dependency graph. Nodes have an average dependency depth of 30 (median of 17), but range anywhere from 0 to 80. We also find a stark difference in dependency depth based on the field of mathematics. For example, entities from textbooks on Lie algebra or Quantum groups had an average depth of $\sim50$, whereas those from books on set theory and logic had an average depth of $\sim6$. This is in-line with the intuition that certain fields of mathematics have more prerequisites, and thus deeper dependency trees.

\subsection{Construction of \datasetname}%
\label{subsec:construction}
Here we give an overview of our dataset construction pipeline. For additional details, see \autoref{app:extraction-pipeline}.

To construct \datasetname, we collect 103 PDFs of math texts from a variety of sources and several fields of mathematics including subfields of algebra, analysis, geometry, number theory, probability, quantum theory, category theory, topology, and others. The math texts cover a total of 87 fields and subfields of mathematics. Following our pipeline (\autoref{fig:creation-pipeline}), we convert the PDFs to MMD (mathematical markdown) files and filter out pages containing metadata, such as indices or glossaries. Then, we run an entity extraction model on the MMDs, which identifies spans of text corresponding to entity types, along with any identifier found in the entity (e.g., ``Example 3.4''). Next, on any extracted entity, we run a reference extractor to identify any references to other entities, objects, or local variables. Lastly, on \textit{definition} entities, we run a name extraction system, which identifies the name(s) of the defined object(s).

After extracting entities and associated metadata, we construct our dependency graph with a relation extraction system (see \autoref{app:extraction-pipeline} for system details). We add ``refers-to'' relations between entity references and their target entities, and between an object references and the entity which defines that object, if present. Finally, we add a ``proves'' relation between a proof entity and the theorem entity it proves.

\subsection{Quality Analysis of \datasetname}%
\label{subsec:data-quality-analysis}
We measure the intrinsic entity and relation extraction quality of the entities in \datasetname.

To evaluate entity extraction quality, we annotate a random sample of $250$ extracted entities from \datasetname, 50 entities of each type, based on whether the entity is valid.\footnote{Our annotators hold, at minimum, undergraduate math degrees.} For an entity to be valid, it must be complete (e.g., not cut off at the start or end, no missing segments due to improper PDF-markdown conversion), and have a correctly labeled type. We find that overall, 90.4\% of the annotated entities are correctly extracted (see \autoref{tab:entity-annotations}).

\jmf{add stuff about precision and recall}
We also annotate 100 randomly sampled references, comprised of 50 object references and 50 entity references. Like before, references are considered correct if they are complete and are labeled with the correct type. We find that 96\% of references were properly extracted. Of the four examples with errors, three were identifiers which were mistaken as entity references, and one was a name mistaken for an object reference.

Finally, we measure the performance of our relation extraction system. For each of the 100 randomly sampled references above, we retrieve 10 candidate targets with LeanSearch and annotate which, if any, are correct. We compute precision, recall, and F1 score of our automated system against the human annotations. Overall, we find an F1 score of 94.8\% (see \autoref{tab:relation-extraction-results}).

\begin{table}[t]
  \centering
  \caption{Intrinsic data quality analysis of \datasetname.}
  \label{tab:entity-annotations}
  \vspace{0.5em}
  \begin{subtable}{0.48\linewidth}
    \centering
    \begin{tabular}[t]{l l l l}
      \toprule
      Ref. type & F1  & Recall & Precision  \\
      \midrule
      Entity ref. & 94.7\% & 100\% & 90\% \\
      Object ref. & 94.9\% & 98\% & 92\% \\
      \midrule
      Overall & 94.8\% & 99\% & 91\% \\
      \bottomrule
    \end{tabular}
    \caption{Performance of our relation extraction system on a human-annotated subset of 100 extracted relations (50 of each type). To measure recall, we annotate gold target entities from among the retrieved candidates; if no selected candidate is an appropriate match, we assume none exists.}
    \label{tab:relation-extraction-results}
  \end{subtable}
  \hfill
  \begin{subtable}{0.48\linewidth}
    \centering
    \begin{tabular}[t]{l l}
      \toprule
      Entity Type & Valid \% \\
      \midrule
      Definition & 98.0\% \\
      Theorem    & 94.0\% \\
      Proof      & 80.0\% \\
      Example    & 86.0\% \\
      Exercise   & 94.0\% \\
      \midrule
      Overall & 90.4\% \\
      \bottomrule
    \end{tabular}
    \caption{Precision of our entity extraction system on a human-annotated subset of 250 examples (50 of each type).}
  \end{subtable}
\end{table}

\section{Autoformalization Metrics for \datasetname}
\label{sec:metrics}

Following previous work, we use the standard autoformalization metric of \textbf{compile rate}.\footnote{We use the software package \texttt{blv} \cite{blv} to measure compile rate, compiled against Lean version v4.24.0.} However, compile rate does not measure faithfulness of the formalization, so we also report \textit{semantic faithfulness} (see \autoref{sec:semantic-faithfulness} for a full discussion).

We report a combined metric of compile rate and semantic faithfulness which we call \textbf{correctness}. A formal statement must both compile and be faithful to the informal statement for it to be considered correct.

\subsection{Semantic Faithfulness Metric}%
\label{sec:semantic-faithfulness}

Here, we discuss semantic faithfulness as a metric, and present experiments showing limitations of prior methods and our improvements. 

Together with syntactic correctness (compilation rate), semantic correctness, or \textit{semantic faithfulness} is a critical metric for autoformalization. Faithfulness measures whether a formal statement has the same mathematical meaning as its informal source. Prior work on semantic faithfulness \cite{luFormalAlignAutomatedAlignment2024,chen2026reformreflectiveautoformalizationprospective, peng2025criticleancriticguidedreinforcementlearning, liu2026assesssemanticstructuralevaluation} focuses on easier domains (olympiad or undergraduate math), and does not measure alignment of definitions. \datasetname~highlights this gap in prior work, as it contains both high-level mathematics and definition entities. 

To assess the quality of semantic faithfulness metrics on graduate-level, \textit{in the wild} mathematics, we introduce \alignmentname, a semantic faithfulness benchmark of syntactically correct 200 statements and definitions from \datasetname. Each informal entity is formalized by \texttt{gpt-oss-120b}, and annotated with a binary label, ``aligned'' or ``misaligned''.\footnote{We selectively sample 200 such entities to better balance the label distribution.} We evaluate prior LLM-as-judge systems, CriticLean \cite{peng2025criticleancriticguidedreinforcementlearning}, as well as
our few-shot LLM-as-judge on ConsistencyCheck \cite{chen2026reformreflectiveautoformalizationprospective} and CriticLeanBench \cite{peng2025criticleancriticguidedreinforcementlearning}, and \alignmentname~(see \autoref{tab:semantic-faithfulness-results}). \footnote{We note that we leave out FormalAlign \cite{luFormalAlignAutomatedAlignment2024} as a baseline as there is no official implementation, and our reimplementation was unable to match the reported results.}


\subsection{Semantic Faithfulness Results}%
\label{sec:semantic-faithfulness-results}
We now analyze the results of our semantic faithfulness experiments (\autoref{tab:semantic-faithfulness-results}). Overall, we find that CriticLean has strong performance on ConsistencyCheck and CriticLeanBench, but a notable drop in accuracy on \alignmentname. In contrast, the strongest LLM-as-judge system (gpt-5.2 with an engineered prompt) drops performance on ConsistencyCheck and CriticLeanBench, but improves by 6\% on \alignmentname~for definitions and 5\% for statements. We suspect this tradeoff is in part due to the selected examples, which were selected from \datasetname~and therefore in-domain for \alignmentname.

However, as gpt-5.2 is both proprietary and expensive to use, we elect to use CriticLean (32B) as our primary semantic faithfulness metric for \datasetname, as it is still well-correlated with human evaluations, is open source, 

\begin{table}
  \centering
  \caption{Results of various methods on several semantic faithfulness benchmarks. ``ReForm prompt'' refers to the faithfulness prompt used by \citet{chen2026reformreflectiveautoformalizationprospective}, and ``Our Prompt'' is an engineered, few-shot prompt optimized for gpt 5.2 with examples taken from \datasetname.}
  \label{tab:semantic-faithfulness-results}
  \vspace{0.5em}
  \begin{tabular}{@{}l c c c c@{}}
    \toprule
    Model & ConsistencyCheck & CriticLeanBench & \alignmentname~Defs. & \alignmentname~Stmts. \\
    \midrule
    \rowcolor{gray!20}
    \multicolumn{5}{c}{\textbf{CriticLean}}                          \\
    CriticLean (14B)             & 81.7     & 84.4    & 68.0 & 56.0  \\
    CriticLean (32B)             & 82.6     & \textbf{86.4}    & 80.0 & 75.0  \\
    \midrule
    \rowcolor{gray!20}
    \multicolumn{5}{c}{\textbf{ReForm Prompt}}     \\
    gpt-oss-120b                 & \textbf{84.9}  & 85.4 & 76.0 & 61.0  \\
    Qwen3 (14B)                  & 80.9  & 82.4 & 70.0 & 57.0  \\
    Qwen3 (32B)                  & 74.6  & 71.5 & 74.0 & 67.0  \\
    \midrule
    \rowcolor{gray!20}
    \multicolumn{5}{c}{\textbf{Our Prompt (Few-shot)}}     \\
    gpt-oss-120b           & 81.1   & 83.8 & 78.0 & 65.0 \\
    Qwen3 (14B)            & 71.3   & 59.4 & 68.0 & 71.0 \\
    Qwen3 (32B)            & 80.5   & 72.4 & 71.0 & 68.0    \\
    gpt-5.2                & 82.1   & 79.0 & \textbf{86.0} & \textbf{80.0} \\
    \bottomrule
  \end{tabular}
\end{table}

\nilay{Add more LLMs to round out experiments in each column.}
\nilay{FormalAlign: discuss how we couldn't reimplement from their exact work}

\section{Autoformalization Experiments}%
\label{sec:autoformalization-experiments}

In this section, we present autoformalization experiments on \datasetname~with several strong baseline methods. We first detail the various models and settings, followed by an analysis of the results.

\subsection{Dataset Splits}
\label{subsec:dataset-splits}

\datasetname~is naturally divided into five splits corresponding to each entity type: \textit{definitions, theorems, exercises, examples}, and \textit{proofs}. However, exercises, examples, and theorems are formalized similarly: that is, as a theorem statement in Lean 4. Thus, we report results on a joint ``statement'' split comprised of all exercises, examples, and theorems. Note that only a correct \textit{statement} is required, not the full proof. We leave the study of proof and full-theorem formalization for future work.

Additionally, we analyze several variables pertinent to high-level, \textit{in the wild} autoformalization. First, we investigate whether an item already existing in Mathlib affects a system's autoformalization performance. It is likely that LLMs have seen Mathlib in its training data. Additionally, Mathlib is constructed bottom-up, meaning entities lower-level fields are more likely to already be formalized. These reasons suggest that existing items will have a higher average correctness. Secondly, we determine to what extent \textit{max dependency depth} affects correctness, as well as \textit{dependency mass} (i.e., the total number of dependencies in the dependency tree).

\paragraph{\hardname} We select a subset of $\sim700$ examples with the deepest dependency trees, which we call \hardname. This set primarily consists of examples from more advanced graduate-level mathematics, such as Algebraic Geometry and Lie Theory, and spans 14 subjects. We also note that \hardname~consists of both statements ($\sim600$) and definitions ($\sim100$).

\paragraph{Open Split} Approximately 30\% of entities in \datasetname~are extracted from Springer textbooks. To comply with copyright law, we release code to generate \datasetname~in full, which can be run by those who have access to Springer textbooks. However, to facilitate access to \datasetname, we release an open split of the remaining 70\% of the data publicly.\footnote{\url{https://huggingface.co/datasets/MathAtlas/MathAtlas}} All examples from \hardname~are in the open split. Our results on this open subset are reported in the Appendix in \autoref{tab:open-results}.


\subsection{Evaluated Methods}
\label{subsec:evaluated-methods}

We measure the performance variety of autoformalization baselines, including fine-tuned methods and prompting methods. We evaluate several strong models with zero and few-shot prompting. We test zero-shot and few-shot prompting with \texttt{gpt-oss-120B} and \texttt{gpt-oss-20b} \cite{openai2025gptoss120bgptoss20bmodel} (see \autoref{app:prompts} for full prompts).  We supply the target entity text as input, and the expected output is a syntactically and semantically correct formal representation in Lean 4 (see \autoref{sec:metrics}). This approach has been extensively studied in prior work, especially on benchmarks such as miniF2F \cite{zhengMiniF2FCrosssystemBenchmark2021} and ProofNet \cite{azerbayevProofNetBenchmarkAutoformalizing}. We perform an ablation, evaluating zero-shot prompting, few-shot with randomly selected examples from LeanWorkbook \cite{ying2024lean}, adding an engineered prompt, and using annotated, graduate-level examples. We use the zero-shot prompt from \citet{zhang-etal-2025-autoformalization}. Our engineered prompt is more detailed in its instructions. Importantly, it instructs the model to synthesize background theory when necessary, and allows for multiple formal output statements as is sometimes required in \datasetname.

On statements, we also test fine-tuned methods: Herald Translator \cite{gao2025heraldnaturallanguageannotated}, Goedel Formalizer v2 \cite{lin2025goedelproverv2scalingformaltheorem}, Kimina Autoformalizer \cite{wang2025kiminaproverpreviewlargeformal}, Atlas Translator \cite{liu2025atlasautoformalizingtheoremslifting}, and ReForm \cite{chen2026reformreflectiveautoformalizationprospective}. However, for definitions, we do not include any fine-tuned methods, as our experiments revealed autoformalization methods for theorems were unable to generalize to definitions in this domain.\footnote{We are unable to evaluate some prior definition autoformalization methods such as ARIA \cite{wang2025ariaagentretrievaliterative} or LoC-Decomp \cite{shi2026locdecomp} due to unreleased code.}

\paragraph{Added Local Context}
As discussed in \autoref{subsec:dataset-structure}, some entities in \datasetname~require context to define variables, entities, or otherwise elaborate a mathematical setting. We evaluate a simple baseline of including the prior 500 tokens in the context window of the LLM. For this setting, we test only the few-shot prompted systems, as this is out-of-domain for the fine-tuned models.


\subsection{Results}%
\label{sec:autoformalization-results}

\begin{table}
  \centering
  \caption{Results for autoformalization experiments on \datasetname. ``Compiles'' measures the percentage of examples which successfully compile. ``Faithful'' measures what percent of the compiling theorems are faithful. ``Correct'' indicates the overall percentage of statements which compile and are faithful.}
  \label{tab:baseline-results}
  \begin{subtable}[t]{0.4\linewidth}
    \centering
    \small
    \begin{tabular}{@{}llll@{}}
      \toprule
      Model & Compiles & Faithful & Correct \\
      \midrule
      \rowcolor{gray!20}
      \multicolumn{4}{c}{\textbf{Prompted}} \\
      gpt-oss-20b (zs) & 10.5\% & 20.9\% & 2.2\% \\
      \hspace{0.5em}+few-shot & 18.1\% & 41.4\% & 7.5\% \\
      \hspace{0.5em}+tuned prompt & 17.1\% & 47.4\% & 8.1\% \\
      \hspace{0.5em}+tuned exs. & 20.8\% & 56.3\% & 11.7\%\\
      \hspace{0.5em}+context & 16.1\% & 55.9\% & 9.0\% \\
      \midrule
      gpt-oss-120b (zs)  & 13.4\% & 49.3\% & 6.6\% \\
      \hspace{0.5em}+few-shot & 24.2\% & 57.0\% & 13.8\% \\
      \hspace{0.5em}+tuned prompt & 26.0\% & 58.8\% & 15.3\% \\
      \hspace{0.5em}+tuned exs. & \textbf{28.5\%} & 58.5\% &  \textbf{16.7\%} \\
      \hspace{0.5em}+local context & 25.8\% & \textbf{58.9\%} & 15.2\% \\
      \bottomrule
    \end{tabular}
    \caption{Autoformalization results for definitions. We do not evaluate fine-tuned methods for definitions because there are no prior-work models available for definitions. The best performance in each category and column are bolded.}
  \end{subtable}
  \hspace{3em}
  \begin{subtable}[t]{0.45\linewidth}
    \centering
    \small
   \begin{tabular}{@{}llll@{}}
      \toprule
      Model & Compiles & Faithful & Correct \\
      \midrule
      \rowcolor{gray!20}
      \multicolumn{4}{c}{\textbf{Fine-tuned}} \\
      Herald 7B & 11.8\% & 12.7\% & 1.5\% \\
      Kimina 7B & \textbf{27.3\%} & 8.4\% & 2.3\%\\
      ATLAS-L 8B & 21.4\% & 16.4\% & 3.5\% \\
      Goedel 8B & 20.3\% & 34.9\% & 7.1\% \\
      Goedel 32B & 23.2\% & 30.6\% & 7.1\% \\
      ReForm 8B & 19.0\% & \textbf{48.4\%} & \textbf{9.8\%} \\
      \midrule
      \rowcolor{gray!20}
      \multicolumn{4}{c}{\textbf{Prompted}} \\
      gpt-oss-20b (zs) & 9.9\% & 42.4\% & 4.2\% \\
      \hspace{0.5em}+few-shot & 16.5\% & 35.8\% & 5.9\% \\
      \hspace{0.5em}+tuned prompt & 16.8\% & 32.7\% & 5.5\% \\
      \hspace{0.5em}+tuned exs. & 19.1\% & 35.1\% & 6.7\% \\
      \hspace{0.5em}+local context & 18.6\% & 36.0\% & 6.7\% \\
      \midrule
      gpt-oss-120b (zs) & 14.3\% & \textbf{47.6\%} & 6.8\% \\
      \hspace{0.5em}+few-shot & 16.8\% & 43.5\% & 7.3\% \\
      \hspace{0.5em}+tuned prompt & 17.1\% & 43.9\% & 7.5\% \\
      \hspace{0.5em}+tuned exs. & \textbf{21.7\%} & 35.9\% & \textbf{7.8\%} \\
      \hspace{0.5em}+local context & 20.0\% & 34.5\% & 6.9\% \\
      \bottomrule
    \end{tabular} 
    \caption{Autoformalization results for statements (theorems, examples, and exercises). The best performance in each category and column are bolded.}
  \end{subtable}
\end{table}

In this section we analyze the results of our autoformalization experiments on \datasetname.

Overall, we find \datasetname~to be very challenging. On the \textit{statement} split, our strongest baseline is ReForm 8B, which achieved an overall correctness of 9.8\% (see \autoref{tab:baseline-results}). We note the disparity between compilation rate and correctness, especially in fine-tuned models: Kimina 7B, for example, achieves 27.3\% compilation rate, but just 2.3\% of formalizations compile and are faithful. On the \textit{definition} split, our strongest system is \texttt{gpt-oss-120b}, with 16.7\% correctness, likewise with a significant 11.8\% gap between compilation rate and overall correctness. As shown in \autoref{tab:hard-results}, \hardname~poses a significant challenge, with our strongest model achieving only 2.6\% correctness.

\paragraph{Local Context} Many items in \datasetname~require local context to correctly solve. However, including 300 tokens of local context seems to confuse prompted systems, and the input decreases system performance from 20.3\% to 17.4\% on definitions, and 10.5\% to 8.6\% on statements for the best model (see \autoref{tab:baseline-results}). This drop in performance indicates further research is necessary on incorporating local context into autoformalization methods.
\begin{wrapfigure}[15]{r}{0.45\textwidth}
\begin{center}
\includegraphics[width=0.9\linewidth]{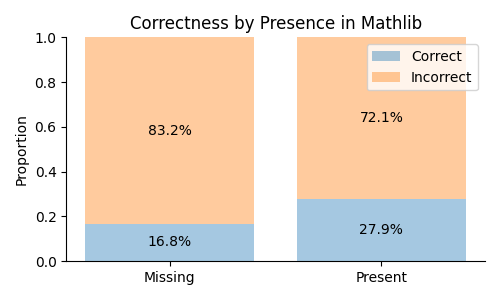}
\caption{Correctness of formalized entities from our best models compared with presence in Mathlib. Entities which our retrieval system (LeanSearch \cite{gao2025semanticsearchenginemathlib4}) found in Mathlib are significantly more likely to be correctly formalized than otherwise.}
\label{fig:correctness-vs-mathlib}
\end{center}
\end{wrapfigure}



\begin{figure}
  \begin{subfigure}[t]{0.48\linewidth}
    \centering
    \includegraphics[width=\linewidth]{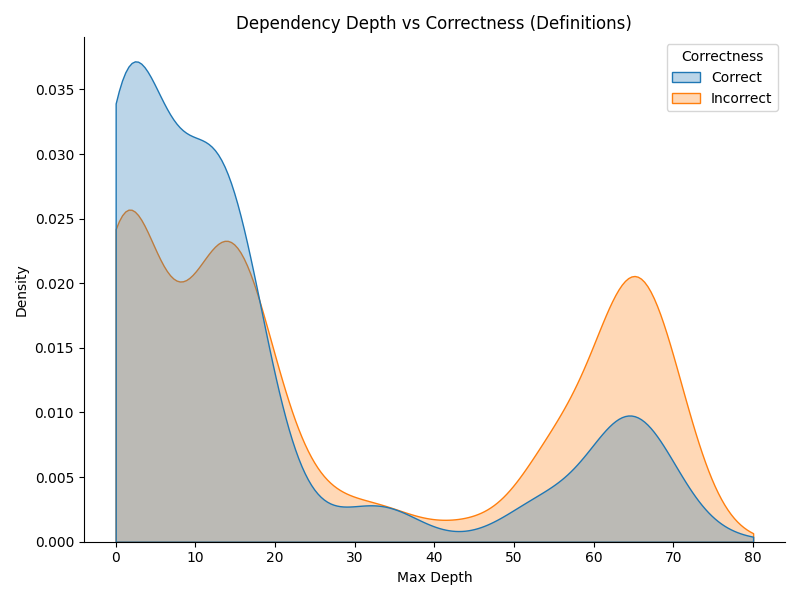}
    \caption{Correctness versus dependency depth of our best performing system for definitions (\texttt{gpt-oss-120b}).}
    \label{fig:correctness-vs-depth-defs}
  \end{subfigure}
  \hfill
  \begin{subfigure}[t]{0.48\linewidth}
    \centering
    \includegraphics[width=\linewidth]{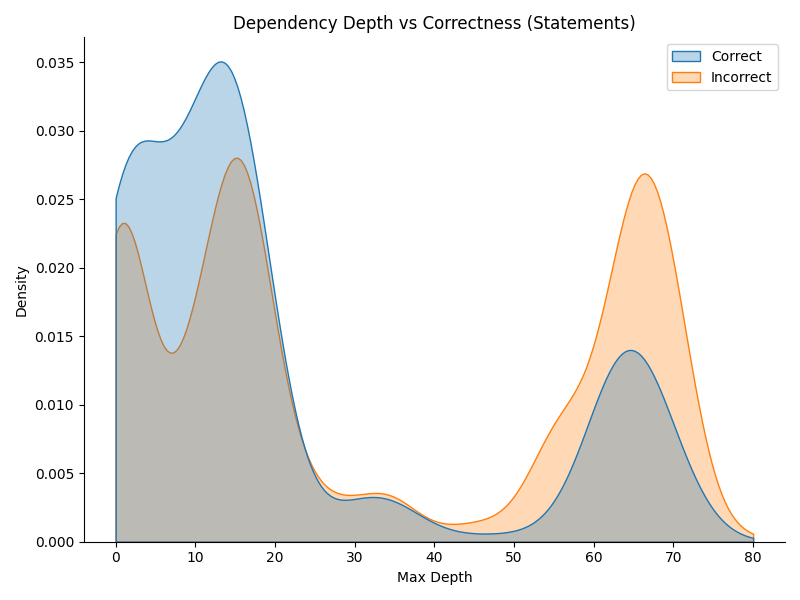}
    \caption{Correctness versus dependency depth of our best performing system for statements (ReForm 8B)}
    \label{fig:correctness-vs-depth-stmts}
  \end{subfigure}
  \caption{Correctness of our strongest models compared with dependency depth on examples in \datasetname.}
    \label{fig:correctness-vs-depth}
\end{figure}

\paragraph{Object Dependencies}
Next, we compare correctness versus max dependency depth. In \autoref{fig:correctness-vs-depth}, we notice a clear trend, showing that our best systems is less likely to correctly formalize deeper targets. We compute a two-sample Kolmogorov–Smirnov, and find significant difference between the max depth distributions for correct and incorrect examples ($p < 0.001$).

\paragraph{Existence in Mathlib}
We compare correctness of definitions which are and aren't grounded in Mathlib (see \autoref{fig:correctness-vs-mathlib}). 
We find that 27.9\% of definitions which our linker grounded in Mathlib are correctly formalized, whereas only 16.8\% of missing definitions are correct, statistically significant difference (chi-square $p < 0.001$). This supports intuition, as Mathlib is likely to be in the training data for modern LLMs like \texttt{gpt-oss-120b}, which may result in inflated autoformalization performance on those entities.

\begin{table}[t]
    \centering
    \caption{Performance of few-shot prompted models on \hardname, which contains both definitions and statements.}
    \label{tab:hard-results}
    \begin{tabular} {llll}
    \toprule
       Model & Compiling & Faithful & Correct \\
    \midrule
        gpt-oss-20b  & 14.3\% & 14.7\% & 2.1\% \\
        gpt-oss-120b & 14.6\% & 17.8\% & 2.6\% \\
    \bottomrule
    \end{tabular}
\end{table}



\section{Related Works}
\label{sec:related-works}

\paragraph{LLMs for Autoformalization}
Large language models (LLMs) have made significant advancements in autoformalization. Early work with LLMs such as in \citet{wuAutoformalizationLargeLanguage2022} used naive approaches, but showed promise in the methodology. Further refinements with fine-tuning on various data made significant progress \citep{gao2025heraldnaturallanguageannotated,lin2025goedelproverv2scalingformaltheorem,wu2025stepfunformalizerunlockingautoformalizationpotential,liu2025atlasautoformalizingtheoremslifting, azerbayevLlemmaOpenLanguage2023} but, as we show in this paper, face similar issues in more difficult settings. More recently, there has been work on breaking down targets either by decomposing into subproblems \citep{shi2026locdecomp} or retrieving dependencies first \citep{wang2025ariaagentretrievaliterative, min2026divide}. These methods have encouraging results in harder settings \citep{jiang2025fateformalbenchmarkseries}, but use proprietary and expensive models and are token-inefficient. ProofFlow \citep{cabralProofFlowDependencyGraph2025} introduces a method for formalizing natural language proofs, as well as ProofScore, a metric for measuring faithfulness to the natural language.

\paragraph{Autoformalization Benchmarks}
Prior autoformalization benchmarks have largely focused on formalizing theorem statements \cite{zhengMiniF2FCrosssystemBenchmark2021, azerbayevProofNetBenchmarkAutoformalizing}, primarily on olympiad or undergraduate mathematics. LeanEuclid \citep{murphy2024leaneuclid} is a benchmark for autoformalizing geometry from Euclid's \textit{Elements} and the UniGeo dataset \citep{chen2022unigeounifyinggeometrylogical}. \citet{zhang-etal-2025-autoformalization} present two definition-specific benchmark of 100 definitions taken from arXiv and Wikipedia, and show proprietary LLMs have low performance. The FATE-H and FATE-X benchmarks \citep{jiang2025fateformalbenchmarkseries} consist of 100 problems from commutative and abstract algebra, and pose a significant challenge compared to previous work. In contrast to the FATE benchmarks, \datasetname~is much larger, contains multiple entity types, is more varied in topic, andwincludes local and long-range dependencies which facilitate the study of isolated system components, such as retrieval or synthesis.

\paragraph{Autoformalization Metrics}
Correctness in autoformalization is measured with two metrics: compilation and semantic faithfulness. The former is simple to measure with the language's compiler, or specialized tools such as Kimina Lean Server \cite{santos2025kiminaleanservertechnical} or blv \cite{blv}.

More difficult to measure is faithfulness, which measures whether the generated formal language code is semantically equivalent to the input informal statement. FormalAlign \citep{luFormalAlignAutomatedAlignment2024} introduces a methodology for fine-tuning an alignment model for statements. However, the methodology does not scale well to more difficult domains, and fails to capture the types of mistakes made by more advanced autoformalization models.\footnote{No implementation or model weights have been released as of this paper. Our analysis of these results is based on our re-implementation, which failed to match reported numbers.} For benchmarks with gold formal labels, BEq \citep{liu2025rethinking} judges semantic equivalence between two formal statements based on definitional equivalence, and TransTED \citep{liu2026assesssemanticstructuralevaluation} presents a tree-based similarity metric. ProofScore \citep{cabralProofFlowDependencyGraph2025} as previously mentioned is a faithfulness metric specifically for proofs. Using an LLM-as-judge approach has been common for measuring semantic faithfulness \citep{zhangGoldStandardsEpistemic2025}, and has shown promise as a scalable metric. ConsistencyCheck \citep{chen2026reformreflectiveautoformalizationprospective} is a benchmark for semantic faithfulness metrics, but as we show in \autoref{sec:semantic-faithfulness-results}, high performance on ConsistencyCheck does not necessarily imply high performance in harder domains such as \alignmentname.

\section{Conclusion}%
\label{sec:conclusion}
%
%

In conclusion, we introduce \datasetname, a large-scale benchmark for graduate-level autoformalization of mathematics in the wild. Through extensive experiments, we show that that \datasetname~is very challenging for existing autoformalization systems.

In contrast to prior benchmarks, \datasetname~with a dependency-graph structure which identifies prerequisite theory for each entity. Our analysis shows that entities with deeper dependency trees are substantially less likely to be formalized correctly. To study this difficulty, we construct \hardname, a subset of 700 entities with the deepest dependency trees in \datasetname, on which our strongest system achieves only 2.6\% correctness. We additionally observe that entities successfully grounded to existing formalizations in Mathlib are significantly more likely to be formalized correctly.

Finally, we introduce \alignmentname, a benchmark for evaluating semantic faithfulness on both mathematical statements and definitions for \datasetname. Experiments on \alignmentname~show that trained metrics such as CriticLean, as well as strong LLM-as-judge systems, have clear room for improvement in faithfulness evaluation.

We release \datasetname, \alignmentname, and our experimental code to support future research for \textit{in the wild} autoformalization.

\section*{Limitations}%
\label{sec:limitations}
\datasetname~primarily deals with graduate-level mathematics, and leaves more difficult domains such as research mathematics for future work. Certain textbooks are also omitted from \datasetname~due to copyright concerns, which can potentially bias data. 

\section*{Acknowledgements}%
\label{sec:acknowledgements}
This work was supported in part by the National Science Foundation under the Artificial Intelligence, Formal Methods, and Mathematical Reasoning (AIMing) NSF grant, award $\#2523479$. This work used resources available through the National Research Platform (NRP) at the University of California, San Diego. NRP has been developed, and is supported in part, by funding from National Science Foundation, from awards 1730158, 1540112, 1541349, 1826967, 2112167, 2100237, and 2120019, as well as additional funding from community partner.

\bibliography{refs}

@misc{Scholze_Buzzard_2020,
  title        = {Liquid Tensor Experiment},
  author       = {Scholze, Peter and Buzzard, Kevin},
  year         = 2020,
  month        = dec,
  day          = 5,
  journal      = {Xena Project},
  url          = {https://xenaproject.wordpress.com/2020/12/05/liquid-tensor-experiment/}
}

@misc{Tao_2023,
  title        = {Formalizing the proof of PFR in Lean4 using Blueprint: a short tour},
  author       = {Tao, Terrance},
  year         = 2023,
  month        = nov,
  journal      = {What’s new},
  url          = {https://terrytao.wordpress.com/2023/11/18/formalizing-the-proof-of-pfr-in-lean4-using-blueprint-a-short-tour}
}

@online{agrawalMathematicsFormalisationAssistant2022,
  title        = {Towards a {{Mathematics Formalisation Assistant}} Using {{Large Language Models}}},
  author       = {Agrawal, Ayush and Gadgil, Siddhartha and Goyal, Navin and Narayanan, Ashvni and Tadipatri, Anand},
  url          = {http://arxiv.org/abs/2211.07524},
  urldate      = {2023-01-10},
  date         = {2022-11-14},
  eprint       = {2211.07524},
  eprinttype   = {arXiv},
  eprintclass  = {cs},
  langid       = {english},
  pubstate     = {prepublished}
}

@online{wuAutoformalizationLargeLanguage2022,
  title        = {Autoformalization with {{Large Language Models}}},
  author       = {Wu, Yuhuai and Jiang, Albert Q. and Li, Wenda and Rabe, Markus N. and Staats, Charles and Jamnik, Mateja and Szegedy, Christian},
  url          = {http://arxiv.org/abs/2205.12615},
  urldate      = {2023-01-10},
  date         = {2022-05-25},
  eprint       = {2205.12615},
  eprinttype   = {arXiv},
  eprintclass  = {cs},
  langid       = {english},
  pubstate     = {prepublished}
}

@article{gadgilAutomatingFormalisationTheorem,
  title        = {Towards Automating Formalisation of Theorem Statements Using Large Language Models},
  author       = {Gadgil, Siddhartha and Tadipatri, Anand Rao and Agrawal, Ayush and Narayanan, Ashvni and Goyal, Navin},
  langid       = {english}
}

@article{azerbayevProofNetBenchmarkAutoformalizing,
  title        = {{{ProofNet}}: {{A Benchmark}} for {{Autoformalizing}} and {{Formally Proving Undergraduate-Level Mathematics Problems}}},
  author       = {Azerbayev, Zhangir and Piotrowski, Bartosz and Avigad, Jeremy},
  journaltitle = {NeurIPS 2022},
  langid       = {english}
}

@online{ExplorationNeuralMachine,
  title        = {Exploration of Neural Machine Translation in Autoformalization of Mathematics in {{Mizar}} | {{Proceedings}} of the 9th {{ACM SIGPLAN International Conference}} on {{Certified Programs}} and {{Proofs}}},
  doi          = {10.1145/3372885.3373827},
  url          = {https://dl.acm.org/doi/10.1145/3372885.3373827},
  urldate      = {2024-06-04},
  langid       = {english},
  organization = {ACM Conferences}
}

@online{zhengMiniF2FCrosssystemBenchmark2021,
  title        = {{{MiniF2F}}: A Cross-System Benchmark for Formal {{Olympiad-level}} Mathematics},
  shorttitle   = {{MiniF2F}},
  author       = {Zheng, Kunhao and Han, Jesse Michael and Polu, Stanislas},
  url          = {https://arxiv.org/abs/2109.00110v2},
  urldate      = {2024-06-05},
  date         = {2021-08-31},
  langid       = {english},
  organization = {arXiv.org}
}

@article{jiang2023multilingual,
  title        = {Multilingual Mathematical Autoformalization},
  author       = {Albert Q. Jiang and Wenda Li and Mateja Jamnik},
  year         = 2023,
  journal      = {arXiv preprint arXiv: 2311.03755}
}

@article{ying2024lean,
  title        = {Lean Workbook: A large-scale Lean problem set formalized from natural language math problems},
  author       = {Huaiyuan Ying and Zijian Wu and Yihan Geng and Jiayu Wang and Dahua Lin and Kai Chen},
  year         = 2024,
  journal      = {arXiv preprint arXiv: 2406.03847}
}

@misc{blecher2023nougat,
  title        = {Nougat: Neural Optical Understanding for Academic Documents},
  author       = {Lukas Blecher and Guillem Cucurull and Thomas Scialom and Robert Stojnic},
  year         = 2023,
  eprint       = {2308.13418},
  archiveprefix = {arXiv},
  primaryclass = {cs.LG}
}

@inproceedings{ThemathlibCommunity2020,
  title        = {The lean mathematical library},
  author       = {The mathlib Community},
  year         = 2020,
  month        = jan,
  booktitle    = {Proceedings of the 9th ACM SIGPLAN International Conference on Certified Programs and Proofs},
  publisher    = {ACM},
  series       = {POPL ’20},
  doi          = {10.1145/3372885.3373824},
  url          = {http://dx.doi.org/10.1145/3372885.3373824},
  collection   = {POPL ’20}
}

@inproceedings{liu2025rethinking,
  title        = {Rethinking and improving autoformalization: towards a faithful metric and a Dependency Retrieval-based approach},
  author       = {Qi Liu and Xinhao Zheng and Xudong Lu and Qinxiang Cao and Junchi Yan},
  year         = 2025,
  booktitle    = {The Thirteenth International Conference on Learning Representations},
  url          = {https://openreview.net/forum?id=hUb2At2DsQ}
}

@software{Kontorovich_Prime_Number_Theorem_2024,
  title        = {{Prime Number Theorem and More}},
  author       = {Kontorovich, Alex and Tao, Terence},
  year         = 2024,
  month        = jan,
  url          = {https://github.com/AlexKontorovich/PrimeNumberTheoremAnd},
  version      = {0.1.0}
}

@misc{gao2025heraldnaturallanguageannotated,
  title        = {Herald: A Natural Language Annotated Lean 4 Dataset},
  author       = {Guoxiong Gao and Yutong Wang and Jiedong Jiang and Qi Gao and Zihan Qin and Tianyi Xu and Bin Dong},
  year         = 2025,
  url          = {https://arxiv.org/abs/2410.10878},
  eprint       = {2410.10878},
  archiveprefix = {arXiv},
  primaryclass = {cs.CL}
}

@misc{liu2025atlasautoformalizingtheoremslifting,
  title        = {ATLAS: Autoformalizing Theorems through Lifting, Augmentation, and Synthesis of Data},
  author       = {Xiaoyang Liu and Kangjie Bao and Jiashuo Zhang and Yunqi Liu and Yu Chen and Yuntian Liu and Yang Jiao and Tao Luo},
  year         = 2025,
  url          = {https://arxiv.org/abs/2502.05567},
  eprint       = {2502.05567},
  archiveprefix = {arXiv},
  primaryclass = {cs.CL}
}

@inproceedings{zhang-etal-2025-autoformalization,
  title        = {Autoformalization in the Wild: Assessing {LLM}s on Real-World Mathematical Definitions},
  author       = {Zhang, Lan and Valentino, Marco and Freitas, Andre},
  year         = 2025,
  month        = nov,
  booktitle    = {Proceedings of the 2025 Conference on Empirical Methods in Natural Language Processing},
  publisher    = {Association for Computational Linguistics},
  address      = {Suzhou, China},
  pages        = {1720--1738},
  doi          = {10.18653/v1/2025.emnlp-main.90},
  isbn         = {979-8-89176-332-6},
  url          = {https://aclanthology.org/2025.emnlp-main.90/},
  editor       = {Christodoulopoulos, Christos and Chakraborty, Tanmoy and Rose, Carolyn and Peng, Violet}
}

@misc{jiang2025fateformalbenchmarkseries,
  title        = {FATE: A Formal Benchmark Series for Frontier Algebra of Multiple Difficulty Levels},
  author       = {Jiedong Jiang and Wanyi He and Yuefeng Wang and Guoxiong Gao and Yongle Hu and Jingting Wang and Nailing Guan and Peihao Wu and Chunbo Dai and Liang Xiao and Bin Dong},
  year         = 2025,
  url          = {https://arxiv.org/abs/2511.02872},
  eprint       = {2511.02872},
  archiveprefix = {arXiv},
  primaryclass = {cs.LG}
}

@misc{zhangGoldStandardsEpistemic2025,
  title        = {Beyond {{Gold Standards}}: {{Epistemic Ensemble}} of {{LLM Judges}} for {{Formal Mathematical Reasoning}}},
  shorttitle   = {Beyond {{Gold Standards}}},
  author       = {Zhang, Lan and Valentino, Marco and Freitas, Andre},
  year         = 2025,
  month        = jun,
  publisher    = {arXiv},
  number       = {arXiv:2506.10903},
  doi          = {10.48550/arXiv.2506.10903},
  urldate      = {2026-01-22},
  eprint       = {2506.10903},
  primaryclass = {cs},
  archiveprefix = {arXiv}
}

@misc{wang2025ariaagentretrievaliterative,
  title        = {Aria: An Agent For Retrieval and Iterative Auto-Formalization via Dependency Graph},
  author       = {Hanyu Wang and Ruohan Xie and Yutong Wang and Guoxiong Gao and Xintao Yu and Bin Dong},
  year         = 2025,
  url          = {https://arxiv.org/abs/2510.04520},
  eprint       = {2510.04520},
  archiveprefix = {arXiv},
  primaryclass = {cs.AI}
}

@misc{lean4,
  title        = {The Lean 4 Theorem Prover and Programming Language},
  author       = {Moura, Leonardo de and Ullrich, Sebastian},
  year         = 2021,
  booktitle    = {Automated Deduction – CADE 28: 28th International Conference on Automated Deduction, Virtual Event, July 12–15, 2021, Proceedings},
  publisher    = {Springer-Verlag},
  address      = {Berlin, Heidelberg},
  pages        = {625–635},
  doi          = {10.1007/978-3-030-79876-5_37},
  isbn         = {978-3-030-79875-8},
  url          = {https://doi.org/10.1007/978-3-030-79876-5_37},
  numpages     = 11
}

@misc{lin2025goedelproverv2scalingformaltheorem,
  title        = {Goedel-Prover-V2: Scaling Formal Theorem Proving with Scaffolded Data Synthesis and Self-Correction},
  author       = {Yong Lin and Shange Tang and Bohan Lyu and Ziran Yang and Jui-Hui Chung and Haoyu Zhao and Lai Jiang and Yihan Geng and Jiawei Ge and Jingruo Sun and Jiayun Wu and Jiri Gesi and Ximing Lu and David Acuna and Kaiyu Yang and Hongzhou Lin and Yejin Choi and Danqi Chen and Sanjeev Arora and Chi Jin},
  year         = 2025,
  url          = {https://arxiv.org/abs/2508.03613},
  eprint       = {2508.03613},
  archiveprefix = {arXiv},
  primaryclass = {cs.LG}
}

@misc{wu2025stepfunformalizerunlockingautoformalizationpotential,
  title        = {StepFun-Formalizer: Unlocking the Autoformalization Potential of LLMs through Knowledge-Reasoning Fusion},
  author       = {Yutong Wu and Di Huang and Ruosi Wan and Yue Peng and Shijie Shang and Chenrui Cao and Lei Qi and Rui Zhang and Zidong Du and Jie Yan and Xing Hu},
  year         = 2025,
  url          = {https://arxiv.org/abs/2508.04440},
  eprint       = {2508.04440},
  archiveprefix = {arXiv},
  primaryclass = {cs.CL}
}

@misc{azerbayevLlemmaOpenLanguage2023,
  title        = {Llemma: {{An Open Language Model For Mathematics}}},
  shorttitle   = {Llemma},
  author       = {Azerbayev, Zhangir and Schoelkopf, Hailey and Paster, Keiran and Santos, Marco Dos and McAleer, Stephen and Jiang, Albert Q. and Deng, Jia and Biderman, Stella and Welleck, Sean},
  year         = 2023,
  month        = oct,
  publisher    = {arXiv},
  number       = {arXiv:2310.10631},
  doi          = {10.48550/arXiv.2310.10631},
  urldate      = {2023-10-18},
  eprint       = {2310.10631},
  primaryclass = {cs},
  archiveprefix = {arXiv}
}

@inproceedings{min2026divide,
  title        = {Divide and Abstract: Autoformalization via Decomposition and Abstraction Learning},
  author       = {Marcus J. Min and Yeqi Gao and Wilson Sy and Zhaoyu Li and Xujie Si and Osbert Bastani},
  year         = 2026,
  booktitle    = {The Fourteenth International Conference on Learning Representations},
  url          = {https://openreview.net/forum?id=NjgaeXNit3}
}

@inproceedings{shi2026locdecomp,
  title        = {LoC-Decomp: {LLM} Autoformalization via Logical Concept Decomposition and Iterative Feedback Correction},
  author       = {Jiangze Shi and Zhiwei Zhang and Baoquan Ma and Shuai Zhao and Ye Yuan and Guoren Wang},
  year         = 2026,
  booktitle    = {The Fourteenth International Conference on Learning Representations},
  url          = {https://openreview.net/forum?id=0KFQ4F9YEH}
}

@inproceedings{murphy2024leaneuclid,
  title        = {Autoformalizing {Euclidean} Geometry},
  author       = {Murphy, Logan and Yang, Kaiyu and Sun, Jialiang and Li, Zhaoyu and Anandkumar, Anima and Si, Xujie},
  year         = 2024,
  booktitle    = {International Conference on Machine Learning (ICML)}
}

@misc{chen2022unigeounifyinggeometrylogical,
      title={UniGeo: Unifying Geometry Logical Reasoning via Reformulating Mathematical Expression}, 
      author={Jiaqi Chen and Tong Li and Jinghui Qin and Pan Lu and Liang Lin and Chongyu Chen and Xiaodan Liang},
      year={2022},
      eprint={2212.02746},
      archivePrefix={arXiv},
      primaryClass={cs.AI},
      url={https://arxiv.org/abs/2212.02746}, 
}

@misc{santos2025kiminaleanservertechnical,
      title={Kimina Lean Server: Technical Report}, 
      author={Marco Dos Santos and Haiming Wang and Hugues de Saxcé and Ran Wang and Mantas Baksys and Mert Unsal and Junqi Liu and Zhengying Liu and Jia Li},
      year={2025},
      eprint={2504.21230},
      archivePrefix={arXiv},
      primaryClass={cs.LO},
      url={https://arxiv.org/abs/2504.21230}, 
}

@misc{luFormalAlignAutomatedAlignment2024,
    title = {{FormalAlign}: {Automated} {Alignment} {Evaluation} for {Autoformalization}},
    shorttitle = {{FormalAlign}},
    url = {http://arxiv.org/abs/2410.10135},
    doi = {10.48550/arXiv.2410.10135},
    urldate = {2026-01-23},
    publisher = {arXiv},
    author = {Lu, Jianqiao and Wan, Yingjia and Huang, Yinya and Xiong, Jing and Liu, Zhengying and Guo, Zhijiang},
    month = oct,
    year = {2024},
    note = {arXiv:2410.10135 [cs]},
}

@misc{cabralProofFlowDependencyGraph2025,
    title = {{ProofFlow}: {A} {Dependency} {Graph} {Approach} to {Faithful} {Proof} {Autoformalization}},
    shorttitle = {{ProofFlow}},
    url = {http://arxiv.org/abs/2510.15981},
    doi = {10.48550/arXiv.2510.15981},
    urldate = {2025-11-01},
    publisher = {arXiv},
    author = {Cabral, Rafael and Do, Tuan Manh and Yu, Xuejun and Tai, Wai Ming and Feng, Zijin and Shen, Xin},
    month = oct,
    year = {2025},
    note = {arXiv:2510.15981 [cs]},
}

@misc{chen2026reformreflectiveautoformalizationprospective,
      title={ReForm: Reflective Autoformalization with Prospective Bounded Sequence Optimization}, 
      author={Guoxin Chen and Jing Wu and Xinjie Chen and Wayne Xin Zhao and Ruihua Song and Chengxi Li and Kai Fan and Dayiheng Liu and Minpeng Liao},
      year={2026},
      eprint={2510.24592},
      archivePrefix={arXiv},
      primaryClass={cs.CL},
      url={https://arxiv.org/abs/2510.24592}, 
}

@misc{openai2025gptoss120bgptoss20bmodel,
      title={gpt-oss-120b \& gpt-oss-20b Model Card}, 
      author={OpenAI and : and Sandhini Agarwal and Lama Ahmad and Jason Ai and Sam Altman and Andy Applebaum and Edwin Arbus and Rahul K. Arora and Yu Bai and Bowen Baker and Haiming Bao and Boaz Barak and Ally Bennett and Tyler Bertao and Nivedita Brett and Eugene Brevdo and Greg Brockman and Sebastien Bubeck and Che Chang and Kai Chen and Mark Chen and Enoch Cheung and Aidan Clark and Dan Cook and Marat Dukhan and Casey Dvorak and Kevin Fives and Vlad Fomenko and Timur Garipov and Kristian Georgiev and Mia Glaese and Tarun Gogineni and Adam Goucher and Lukas Gross and Katia Gil Guzman and John Hallman and Jackie Hehir and Johannes Heidecke and Alec Helyar and Haitang Hu and Romain Huet and Jacob Huh and Saachi Jain and Zach Johnson and Chris Koch and Irina Kofman and Dominik Kundel and Jason Kwon and Volodymyr Kyrylov and Elaine Ya Le and Guillaume Leclerc and James Park Lennon and Scott Lessans and Mario Lezcano-Casado and Yuanzhi Li and Zhuohan Li and Ji Lin and Jordan Liss and Lily and Liu and Jiancheng Liu and Kevin Lu and Chris Lu and Zoran Martinovic and Lindsay McCallum and Josh McGrath and Scott McKinney and Aidan McLaughlin and Song Mei and Steve Mostovoy and Tong Mu and Gideon Myles and Alexander Neitz and Alex Nichol and Jakub Pachocki and Alex Paino and Dana Palmie and Ashley Pantuliano and Giambattista Parascandolo and Jongsoo Park and Leher Pathak and Carolina Paz and Ludovic Peran and Dmitry Pimenov and Michelle Pokrass and Elizabeth Proehl and Huida Qiu and Gaby Raila and Filippo Raso and Hongyu Ren and Kimmy Richardson and David Robinson and Bob Rotsted and Hadi Salman and Suvansh Sanjeev and Max Schwarzer and D. Sculley and Harshit Sikchi and Kendal Simon and Karan Singhal and Yang Song and Dane Stuckey and Zhiqing Sun and Philippe Tillet and Sam Toizer and Foivos Tsimpourlas and Nikhil Vyas and Eric Wallace and Xin Wang and Miles Wang and Olivia Watkins and Kevin Weil and Amy Wendling and Kevin Whinnery and Cedric Whitney and Hannah Wong and Lin Yang and Yu Yang and Michihiro Yasunaga and Kristen Ying and Wojciech Zaremba and Wenting Zhan and Cyril Zhang and Brian Zhang and Eddie Zhang and Shengjia Zhao},
      year={2025},
      eprint={2508.10925},
      archivePrefix={arXiv},
      primaryClass={cs.CL},
      url={https://arxiv.org/abs/2508.10925}, 
}

@misc{wang2025kiminaproverpreviewlargeformal,
      title={Kimina-Prover Preview: Towards Large Formal Reasoning Models with Reinforcement Learning}, 
      author={Haiming Wang and Mert Unsal and Xiaohan Lin and Mantas Baksys and Junqi Liu and Marco Dos Santos and Flood Sung and Marina Vinyes and Zhenzhe Ying and Zekai Zhu and Jianqiao Lu and Hugues de Saxcé and Bolton Bailey and Chendong Song and Chenjun Xiao and Dehao Zhang and Ebony Zhang and Frederick Pu and Han Zhu and Jiawei Liu and Jonas Bayer and Julien Michel and Longhui Yu and Léo Dreyfus-Schmidt and Lewis Tunstall and Luigi Pagani and Moreira Machado and Pauline Bourigault and Ran Wang and Stanislas Polu and Thibaut Barroyer and Wen-Ding Li and Yazhe Niu and Yann Fleureau and Yangyang Hu and Zhouliang Yu and Zihan Wang and Zhilin Yang and Zhengying Liu and Jia Li},
      year={2025},
      eprint={2504.11354},
      archivePrefix={arXiv},
      primaryClass={cs.AI},
      url={https://arxiv.org/abs/2504.11354}, 
}

@misc{blv,
	author = {{Nilay Patel}},
	year = {2026},
	month = {mar 11},
	title = {offendo/blv},
	url = {https://github.com/offendo/blv},
	howpublished = {https://github.com/offendo/blv},
}

@misc{peng2025criticleancriticguidedreinforcementlearning,
      title={CriticLean: Critic-Guided Reinforcement Learning for Mathematical Formalization}, 
      author={Zhongyuan Peng and Yifan Yao and Kaijing Ma and Shuyue Guo and Yizhe Li and Yichi Zhang and Chenchen Zhang and Yifan Zhang and Zhouliang Yu and Luming Li and Minghao Liu and Yihang Xia and Jiawei Shen and Yuchen Wu and Yixin Cao and Zhaoxiang Zhang and Wenhao Huang and Jiaheng Liu and Ge Zhang},
      year={2025},
      eprint={2507.06181},
      archivePrefix={arXiv},
      primaryClass={cs.CL},
      url={https://arxiv.org/abs/2507.06181}, 
}

@misc{liu2026assesssemanticstructuralevaluation,
      title={ASSESS: A Semantic and Structural Evaluation Framework for Statement Similarity}, 
      author={Xiaoyang Liu and Tao Zhu and Zineng Dong and Yuntian Liu and Qingfeng Guo and Zhaoxuan Liu and Yu Chen and Tao Luo},
      year={2026},
      eprint={2509.22246},
      archivePrefix={arXiv},
      primaryClass={cs.LG},
      url={https://arxiv.org/abs/2509.22246}, 
}

@inproceedings{kwon2023efficient,
  title={Efficient Memory Management for Large Language Model Serving with PagedAttention},
  author={Woosuk Kwon and Zhuohan Li and Siyuan Zhuang and Ying Sheng and Lianmin Zheng and Cody Hao Yu and Joseph E. Gonzalez and Hao Zhang and Ion Stoica},
  booktitle={Proceedings of the ACM SIGOPS 29th Symposium on Operating Systems Principles},
  year={2023}
}

@misc{chromadb,
	author = {Escriva, Robert and Bashir, Hammad and Isom, Max and Huber, Jeff and {Macronova} and Kedia, Sanket and {itaismith} and VanderHart, Luke and Radhakrishnan, Jai and {tanujnay112} and Azarov, Trayan and Eggers, Ben and Diaz, Kyle and Pei, Liquan and {jasonvigil} and Kim, Drew and Troynikov, Anton and Thomas, Philip I. and P, Rohit and {nicolasgere} and Shahbazian, Gabe and Culver, Evan and Gamble, Cooper and Dash, Dave and Krusinski, TJ and Gu, Weili and {swyx.io} and Keller, Matthew and Him, Really and {naynaly10}},
	year = {2026},
	month = {may 7},
	title = {chroma-core/chroma},
	url = {https://github.com/chroma-core/chroma},
	howpublished = {https://github.com/chroma-core/chroma},
}

@misc{wang2024improvingtextembeddingslarge,
      title={Improving Text Embeddings with Large Language Models}, 
      author={Liang Wang and Nan Yang and Xiaolong Huang and Linjun Yang and Rangan Majumder and Furu Wei},
      year={2024},
      eprint={2401.00368},
      archivePrefix={arXiv},
      primaryClass={cs.CL},
      url={https://arxiv.org/abs/2401.00368}, 
}

@misc{gao2025semanticsearchenginemathlib4,
      title={A Semantic Search Engine for Mathlib4}, 
      author={Guoxiong Gao and Haocheng Ju and Jiedong Jiang and Zihan Qin and Bin Dong},
      year={2025},
      eprint={2403.13310},
      archivePrefix={arXiv},
      primaryClass={cs.IR},
      url={https://arxiv.org/abs/2403.13310}, 
}

\newpage
\appendix

\section{Extraction Pipeline}
\label{app:extraction-pipeline}

\begin{figure}
    \centering
    \includegraphics[width=0.7\linewidth]{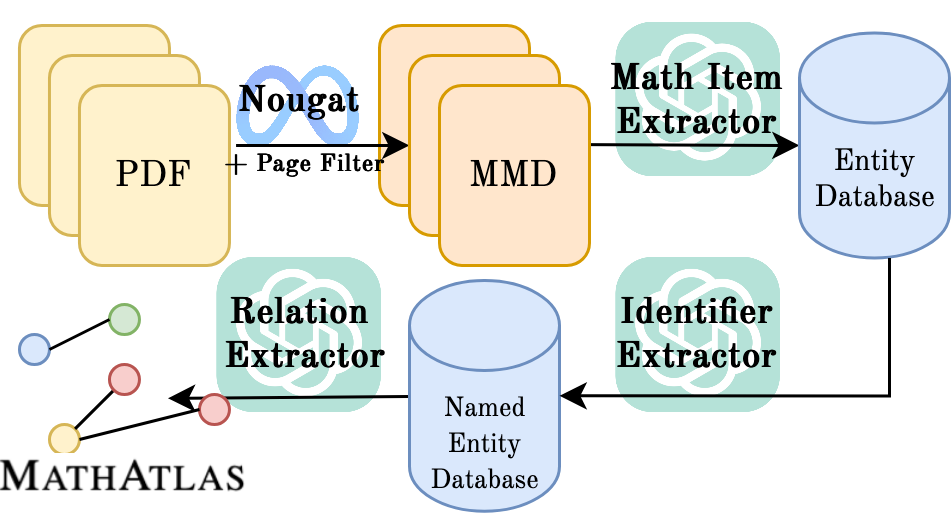}
    \caption{\datasetname~creation pipeline. For visual convenience, we combine the ``name'' and ``reference'' extractors into one step called ``identifier extractor.''}
    \label{fig:creation-pipeline}
\end{figure}
Our extraction pipeline starts with 103 selected textbooks covering a wide range of mathematics. We use Nougat \cite{blecher2023nougat} to convert PDFs into Mathematical Markdown (MMD). We then run our entity extraction system, followed by our reference extraction and name extraction systems, and finally our relation extraction system. We detail each below.

\begin{figure}
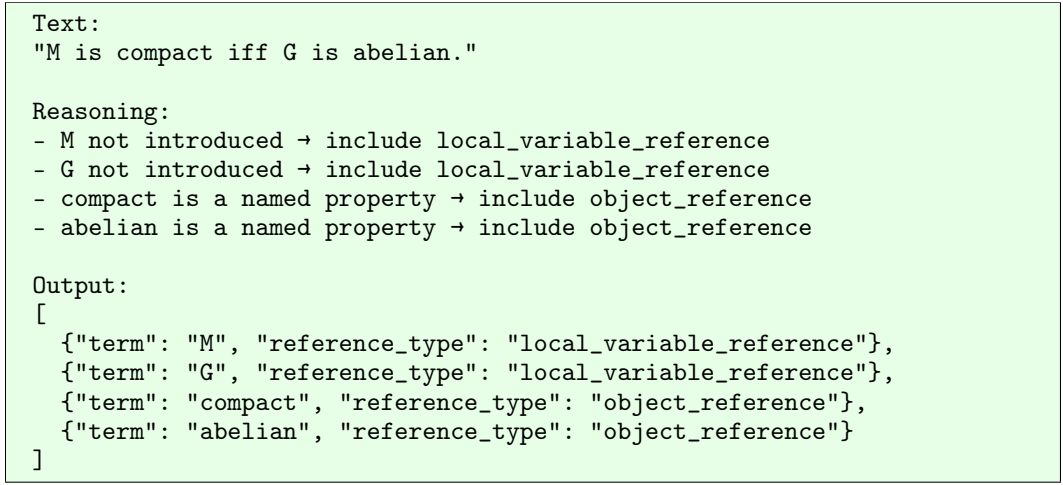

  \begin{mdframed}[backgroundcolor=green!10] 
  \begin{verbatim}
Text:
"M is compact iff G is abelian."

Reasoning:
- M not introduced → include local_variable_reference
- G not introduced → include local_variable_reference
- compact is a named property → include object_reference
- abelian is a named property → include object_reference

Output:
[
  {"term": "M", "reference_type": "local_variable_reference"},
  {"term": "G", "reference_type": "local_variable_reference"},
  {"term": "compact", "reference_type": "object_reference"},
  {"term": "abelian", "reference_type": "object_reference"}
]
\end{verbatim}
  \end{mdframed}
  \caption{Example of reference extraction system input and output.}
\end{figure}

\paragraph{Entity Extraction} Our entity extraction system is a few-shot, prompt-engineered \texttt{gpt-oss-120b}. We prompt the model to generate an array of JSON objects with keys ``type'' indicating the entity's type, optionally an ``identifier'', which are spans such as Definition 3.9 or Theorem 15, and ``text'', which is the actual text of the entity. We enforce this structure with a JSON schema using structured decoding, a temperature of 0.3, and otherwise default sampling parameters.

We then run our reference extractor system, likewise a prompted \textit{gpt-oss-120b} model. The reference extractor takes as input an extracted entity, and outputs a list of JSON objects with keys ``term,'' indicating the text of the reference, and ``reference\_type,'' which is one of \texttt{object\_reference}, \texttt{entity\_reference}, or \texttt{local\_variable\_reference}. This is likewise enforced with JSON schema, and uses a temperature of 0.3.

Then, we run the name extractor, also a prompted \texttt{gpt-oss-120b}, prompted to output the names of objects defined in the entity. The output a JSON array of strings, again enforced with a schema and structured decoding. We likewise use a temperature of 0.3.

\paragraph{Relation Extraction} Each entity is embedded into a ChromaDB instance \cite{chromadb} together with all its metadata, using the Mistral E5 model from \citet{wang2024improvingtextembeddingslarge}. Then, our relation extraction system works in two stages. First, we use the same model to retrieve candidate matches from the entity database. Secondly, we prompt \texttt{gpt-oss-120b} to determine which, if any, is the correct match, accounting for usage and context. If there are multiple matches, we prioritize those from the same textbook, or otherwise those with the highest matching score.

\paragraph{Grounding in Mathlib} We use LeanSearch \cite{gao2025semanticsearchenginemathlib4} to link definitions in \datasetname~to formal definitions in Mathlib. We query the definition text to get 10 candidate matches, and select from the matches using a prompted ``gpt-oss-120b.'' Unlike the authors however, we find that augmenting the query degrades performance, so we skip that step in our final run. We suspect this is because our queries are already fully-formed definitions rather than keywords or phrases. We include the selected matches as well as all 10 candidates for each definition in the released dataset.

\section{Prompts}
\label{app:prompts}

\begin{lstlisting}[breaklines=true,frame=single]
   ## Task
You are given a block of markdown mathematical text.
Your task is to identify and extract all contiguous snippets that correspond to any of the following mathematical entity types:

- Definition
- Theorem (this includes propositions, lemmas, corollaries)
- Example
- Exercise
- Proof

Only extract genuine mathematical content of these types.
Ignore and do NOT extract:

- General exposition or narrative explanation
- Chapter titles, section headings, or indexes
- Transitional text or motivation
- References to earlier or later results without stating them


## What qualifies as an entity

### 1. Definitions (very important)

A definition does NOT require a heading or numbering.

You must extract a block as a definition if its primary purpose is to introduce, name, or precisely describe a mathematical object or concept, even if it appears inline in the text.

Treat a block as a definition if it contains definitional language, including but not limited to:

- "is called"
- "is defined as"
- "we call"
- "is said to be"
- "the ... of ... is"
- "an element ... is"
- an emphasized or italicized term followed by an explanation
- a mathematical object followed by a copular verb ("is", "are") and a precise condition

If a paragraph answers the question "What is X?", it is a definition.

A single paragraph may introduce multiple concepts (e.g., minimum polynomial and conjugate).
If so, extract the entire paragraph as one definition entity.

Unlabeled definitions are mandatory.
Failure to extract them is considered an incorrect output.


### 2. Theorems

Extract any explicitly stated mathematical claim intended to be true, including:

- Theorem
- Proposition
- Lemma
- Corollary

Treat all of these as type "theorem".

The entity must assert a result, not merely mention one.


### 3. Proofs

Extract a proof only when:

- It is explicitly labeled (e.g., "Proof.", "Proof of Proposition 3.2."), and
- It contains actual mathematical reasoning or argument

Special rule:
If a proof appears only as part of an exercise prompt (e.g., "Prove that ...") and contains no mathematical argument beyond the statement itself, do NOT extract a separate proof entity.

Caution:
Be careful to extract a full proof, and do not cut off any information halfway through.


### 4. Examples

Extract examples that demonstrate or instantiate a mathematical concept, whether labeled ("Example 2.3") or clearly signposted in the text.


### 5. Exercises

Extract exercises or problems intended for the reader to solve, whether labeled or clearly formatted as such.


## Output format

For each extracted entity, produce a JSON object with the following keys:

- "type": one of "definition", "theorem", "example", "exercise", "proof" (always lowercase)
- "identifier": a concise label taken verbatim from the text introducing the entity
  (e.g., "Definition 6.9.", "Proposition 6.11.", "Proof.", "Exercise 7(b)")
  If no explicit label, number, or name is present, set this to null.
- "text": the exact text of the entity, copied verbatim.
  Do NOT summarize, paraphrase, normalize spacing, or rewrite mathematics.


## Ordering rules

- Preserve the original order of appearance.
- Each entity must be its own JSON object.
- Do not merge separate entities.
- Do not invent or infer missing text.


## Reasoning step (required)

Before producing the final output:

1. Scan the text systematically from start to end.
2. Actively search for:
   - Explicit labels (Definition, Theorem, Proof, etc.)
   - Implicit definitions using definitional language
3. For each candidate block:
   - Decide whether it is genuine mathematical content
   - Confirm it matches one of the allowed entity types
4. Exclude all non-mathematical exposition.

Only after completing this identification and segmentation should you produce the output.


## Output constraints

- Respond with a JSON list only
- Do NOT wrap the JSON in code fences
- Do NOT include explanations, commentary, or reasoning in the output
- If no valid entities are found, return an empty list: []


## Priority reminder

If a block introduces or defines a mathematical object - even without a heading - it MUST be extracted as a "definition".
\end{lstlisting}
\captionof{figure}{Prompt for entity extraction}

\begin{lstlisting}[breaklines=true,frame=single]
You are given a mathematical definition written in textbook style.

Your task is to extract the names of the mathematical objects, concepts, or properties being defined.

### Rules:
- Return only the names of the items being defined.
- Do not include symbols, formulas, or explanations.
- If multiple items are defined, return all of them.
- If a definition gives multiple synonymous names (e.g. "X (or Y)"), include the most standard or commonly used name.
- Ignore purely notational conventions unless they define a commonly named concept.
- Normalize names to lowercase unless capitalization is standard.
- Output the result as a JSON-style list of strings.

Definition:
{{DEFINITION_TEXT}}

Output format:
["name 1", "name 2", ...]

### Examples:

Example 1:
Definition:
**Definition 0.3.4**.:
1. A _union_ of two sets \(A\) and \(B\) is defined as \[A\cup B\coloneqq\{{x:x\in A\text{{ or }}x\in B\}}.\]
2. An _intersection_ of two sets \(A\) and \(B\) is defined as \[A\cap B\coloneqq\{{x:x\in A\text{{ and }}x\in B\}}.\]
3. A _complement of \(B\) relative to \(A\)_ (or _set-theoretic difference_ of \(A\) and \(B\)) is defined as \[A\setminus B\coloneqq\{{x:x\in A\text{{ and }}x\notin B\}}.\]
4. We say _complement_ of \(B\) and write \(B^{{c}}\) instead of \(A\setminus B\) if the set \(A\) is either the entire universe or if it is the obvious set containing \(B\), and is understood from context.
5. We say sets \(A\) and \(B\) are _disjoint_ if \(A\cap B=\emptyset\)

Output:
["union", "intersection", "set-theoretic difference", "complement", "disjoint"]

Example 2:
Definition:
**Definition 6.1** (\(\mathsf{{RCA}}_{{0}}\)).: An _integral domain_ is a ring \(R\) with no nonzero zero-divisors

Output:
["integral domain"]
\end{lstlisting}
\captionof{figure}{Prompt for name extraction}

\begin{lstlisting}[breaklines=true,frame=single]
You are an expert mathematician and a specialist in formal mathematics, especially Lean 4 and mathlib4.

Your task is to identify all mathematical references that appear in a given mathematical definition, theorem, lemma, or statement.

A reference is any term, symbol, or phrase whose meaning depends on a previously defined mathematical object, entity, or external context.

You must classify each reference into exactly one of the following categories.

--------------------------------------------------
REFERENCE TYPES
--------------------------------------------------

1. "object_reference"

A standard, named mathematical object, structure, property, or concept whose definition exists independently of the current text.

Examples:
- group
- topological space
- metric space
- compact space
- real number
- real vector space
- group homomorphism
- general linear group
- continuous function
- prime number
- finite set

Include:
- Named mathematical structures
- Named mathematical properties if they have formal definitions
- Standard named constructions

Do NOT include:
- Generic words like "element", "set", "function", "number"
- Logical words like "if", "then", "and", "exists"
- Less specific forms of an object. I.e., in the sentence "Let G be an abelian group", you should include "abelian group", NOT "group".

--------------------------------------------------

2. "entity_reference"

A reference to a specific previously defined, labeled, or numbered mathematical item.

Examples:
- Definition 3.4
- Theorem 2.1
- Lemma 5
- Proposition 4.2
- Corollary 1.3
- previous lemma
- above theorem
- following definition

These refer to specific formal entities in the document structure.

--------------------------------------------------

3. "local_variable_reference"

A symbol that refers to a mathematical object but is NOT explicitly introduced in the current text.

A symbol is explicitly introduced if the current text declares it using phrases such as:

- let M be ...
- suppose M is ...
- fix M ...
- given M ...
- where M is ...
- for M a ...
- denote by M ...
- let M := ...
- assume M is ...
- take M ...
- any equivalent explicit declaration

If a symbol is explicitly introduced in the current text, DO NOT include it.

Only include symbols that are used WITHOUT being introduced.

EXAMPLES

Example:

Text:
"Let G be a group. Then G is abelian iff the center of G equals G."

Reasoning:
- G is explicitly introduced -> exclude
- group is a named object -> include
- abelian is a named property -> include
- center is a named construction -> include

Output:
[
  {"term": "group", "reference_type": "object_reference"},
  {"term": "abelian", "reference_type": "object_reference"},
  {"term": "center", "reference_type": "object_reference"}
]

--------------------------------------------------

Example:

Text:
"M is compact iff G is abelian."

Reasoning:
- M not introduced -> include local_variable_reference
- G not introduced -> include local_variable_reference
- compact is a named property -> include object_reference
- abelian is a named property -> include object_reference

Output:
[
  {"term": "M", "reference_type": "local_variable_reference"},
  {"term": "G", "reference_type": "local_variable_reference"},
  {"term": "compact", "reference_type": "object_reference"},
  {"term": "abelian", "reference_type": "object_reference"}
]

--------------------------------------------------

If no references exist, output:

[]
\end{lstlisting}
\captionof{figure}{Prompt for reference extraction}

\begin{lstlisting}[breaklines=true,frame=single]
You are a meticulous expert in Lean 4 and `mathlib4`.

Your task is to act as a "grounding" reasoner for a formalization agent. Your goal is to determine whether a given mathematical concept has a canonical formal definition in `mathlib`, based on a list of search candidates.

--------------------------------------------------
Your Task (Follow These Steps PRECISELY)
--------------------------------------------------

--------------------------------------------------
Step 1: Direct Match Analysis
--------------------------------------------------

- First, look for a **direct, canonical definition** among the candidates.
- A direct match is typically a `class`, `structure`, or `def` whose name is very similar to the concept name.

  Example:
  - Concept: "local ring"
  - Match: `class IsLocalRing`

- If you find a clear, direct match, use that as your primary answer.

--------------------------------------------------
Step 2: Deduction from Usage Patterns
(Only if no direct match is found in Step 1)
--------------------------------------------------

- If no direct match was found in Step 1, your task is to **deduce** the canonical name by finding a **consistent usage pattern** across multiple `theorem` and `instance` candidates.

- **Analyze the signatures:**
  Look for a common identifier that is consistently used as a **type** or **typeclass** across multiple candidates.

- **Example:**
  If you are looking for "CharZero" and the search results include:
  - `instance : CharZero N`
  - `instance : CharZero Z`
  - `theorem my_thm [CharZero R]`

  Then the identifier `CharZero` appears repeatedly as a typeclass. This is overwhelming evidence that the canonical definition is named `CharZero`.

- **Strict Rule:**
  The name you select **must** be an identifier that is explicitly present in the candidate list.

  - Do **not** invent, combine, or guess a new name.
  - If no single, consistent pattern emerges from the candidates, you must conclude that no confident match can be found.

--------------------------------------------------
Step 3: Final Decision
--------------------------------------------------

- Based on your analysis from Step 1 and Step 2, determine the single best candidate for the concept.

- Your answer MUST be a single, valid JSON object with the following keys:

  - `"best_match"`:
    The **list index** of the candidate which precisely matches the search query. If no confident match can be found through either direct matching or inference, the value must be `null`.

  - `"reasoning"`:
    An explanation of why you chose a match, or didn't choose any match.
\end{lstlisting}
\captionof{figure}{Prompt for grounding entities to mathlib. 10 candidate results are fed into this prompt, and the output is an index of the best match and a reasoning.}

\section{Additional Experimental Details}
\label{app:experimental-details}
We use vLLM \cite{kwon2023efficient} as an efficient inference server for all of our dataset construction, autoformalization, and evaluation. If specified, we use the author's recommended sampling parameters. Otherwise, we use a temperature of 0.5, a top-p of 0.95, and otherwise default parameters for all autoformalization experiments. We use deterministic sampling for our semantic faithfulness systems.

All experiments were run on a single H200 GPU. Depending on the model, full-benchmark runs run from 1-10 hours. Checking compilation runs on 10 cpu workers for approximately 5 minutes. Measuring faithfulness runs for $\sim$1-2 hours, depending on the number of successfully compiling examples.

\section{Broader impacts}
\label{app:broader-impacts}
We release \datasetname~with the hope to advance the field of autoformalization, and with it, AI and mathematics in general. This outcome can potentially have larger societal impact as AI becomes more powerful and more widely used. However, it is unlikely that this contribution in particular will have large-scale negative ramifications.

\begin{table}[t]
  \centering
  \caption{Autoformalization results on the ``open'' split of MathAtlas. The results listed here are similar to those in \autoref{tab:baseline-results}, but restricted to a subset of the data.}
  \label{tab:open-results}
  \begin{subtable}[t]{0.4\linewidth}
    \centering
    \small
    \begin{tabular}{@{}llll@{}}
      \toprule
      Model & Compiles & Faithful & Correct \\
      \midrule
      \rowcolor{gray!20}
      \multicolumn{4}{c}{\textbf{Prompted}} \\
      gpt-oss-20b (zs) & 9.5\% & 48.4\% & 4.6\% \\
      \hspace{0.5em}+few-shot & 16.3\% & 47.2\% & 7.7\% \\
      \hspace{0.5em}+tuned prompt & 16.8\%  & 47.6\% & 8.0\% \\
      \hspace{0.5em}+tuned exs. & 19.9\% & 60.3\% & 12.0\%\\
      \midrule
      gpt-oss-120b (zs)  & 13.0\% & 49.2\% & 6.4\% \\
      \hspace{0.5em}+few-shot & 24.2\% & 57.0\% & 13.8\% \\
      \hspace{0.5em}+tuned prompt & 26.0\% & 58.8\% & 15.3\% \\
      \hspace{0.5em}+tuned exs. & \textbf{27.4\%} & \textbf{62.0\%} &  \textbf{17.0\%} \\
      \bottomrule
    \end{tabular}
    \caption{Autoformalization results for definitions. The best performance in each category and column are bolded.}
  \end{subtable}
  \hspace{3em}
  \begin{subtable}[t]{0.45\linewidth}
    \centering
    \small
    \begin{tabular}{@{}llll@{}}
      \toprule
      Model & Compiles & Faithful & Correct \\
      \midrule
      \rowcolor{gray!20}
      \multicolumn{4}{c}{\textbf{Fine-tuned}} \\
      Herald 7B & 11.8\% & 14.4\% & 1.7\% \\
      Kimina 7B & \textbf{27.1\%} & 8.1\% & 2.2\%\\
      ATLAS-L 8B & 20.1\% & 16.9\% & 3.4\% \\
      Goedel 8B & 18.9\% & 37.6\% & 7.1\% \\
      Goedel 32B & 22.3\% & 32.7\% & 7.3\% \\
      ReForm 8B & 18.1\% & \textbf{50.8\%} & \textbf{9.2\%} \\
      \midrule
      \rowcolor{gray!20}
      \multicolumn{4}{c}{\textbf{Prompted}} \\
      gpt-oss-20b (zs) & 9.3\% & 49.5\% & 4.6\% \\
      \hspace{0.5em}+few-shot & 16.5\% & 35.8\% & 5.9\% \\
      \hspace{0.5em}+tuned prompt & 16.8\% & 32.7\% & 5.5\% \\
      \hspace{0.5em}+tuned exs. & 19.1\% & 35.1\% & 6.7\% \\
      \midrule
      gpt-oss-120b (zs) & 15.5\% & 43.2\% & 6.7\% \\
      \hspace{0.5em}+few-shot & 16.3\% & \textbf{44.2\%} & 7.2\% \\
      \hspace{0.5em}+tuned prompt & 17.1\% & 43.9\% & 7.5\% \\
      \hspace{0.5em}+tuned exs. & \textbf{20.7\%} & 38.6\% & \textbf{8.0\%} \\
      \bottomrule
    \end{tabular}
    \caption{Autoformalization results for statements (theorems, examples, and exercises). The best performance in each category and column are bolded.}
  \end{subtable}
\end{table}


\end{document}